\documentclass[conference]{IEEEtran}
\IEEEoverridecommandlockouts
\usepackage{cite}
\usepackage{amsmath,amssymb,amsfonts}
\usepackage{algorithmic}
\usepackage{graphicx}
\usepackage{textcomp}
\usepackage{multirow}
\usepackage[table,xcdraw]{xcolor}

\def\BibTeX{{\rm B\kern-.05em{\sc i\kern-.025em b}\kern-.08em
    T\kern-.1667em\lower.7ex\hbox{E}\kern-.125emX}}
\begin{document}

\title{A Comparative Evaluation of Heart Rate \\Estimation Methods using Face Videos
}

\author{Javier Hernandez-Ortega, Julian Fierrez, Aythami Morales, David Diaz\\
School of Engineering, Universidad Autonoma de Madrid, Spain\\ 
{\small \texttt{\{javier.hernandezo, julian.fierrez, aythami.morales\}@uam.es}}\\ 
}



\maketitle

\begin{abstract}
This paper presents a comparative evaluation of methods for remote heart rate estimation using face videos, i.e., given a video sequence of the face as input, methods to process it to obtain a robust estimation of the subject's heart rate at each moment. Four alternatives from the literature are tested, three based in hand-crafted approaches and one based on deep learning. The methods are compared using RGB videos from the COHFACE database. Experiments show that the learning-based method achieves much better accuracy than the hand-crafted ones. The low error rate achieved by the learning-based model makes possible its application in real scenarios, e.g. in medical or sports environments.
\end{abstract}


\section{Introduction}
\label{sec:intro}


Photoplethysmography (PPG) \cite{allen2007photoplethysmography} is a low-cost technique for measuring the cardiovascular Blood Volume Pulse (BVP) through changes in the amount of light reflected or absorbed by human vessels. This information can be used to estimate parameters such as heart rate, arterial pressure, blood glucose level, or oxygen saturation levels. PPG signals are usually measured with contact sensors placed at the fingertips, the chest, or the feet. This type of contact measurement may be suitable for a clinic environment, but it can be uncomfortable and inconvenient for other application fields such as driver monitoring, sport events, or face antispoofing. Remote PPG (rPPG) consists in applying PPG techniques to face video sequences. These techniques look for changes in the color of the user's face caused by the variation of the oxygen concentration in the blood.

\begin{table*}[t]
\caption{Selection of works that use different types of images to implement rPPG for heart rate extraction or other related tasks like face antispoofing or stress detection. The works evaluated in this paper are highlighted in bold.}
\label{table1}
\begin{center}
\resizebox{\textwidth}{!}{
\begin{tabular}{|l|c|c|c|c|c|}
\hline
Method & Type of Images & Database used & Video Length & Target & Performance \\
\hline\hline
Garbey et al. 2007 \cite{thermal2007garbey} & Thermal & self-collected & 120 secs. & HR Estimation & Accuracy = 99\% \\
\textbf{Verkruysse et al. 2008 \cite{Verkruysse:08}} & RGB & self-collected & 10 secs. & HR \& Respiration Estimation & Qualitative \\
\textbf{Poh et al. 2011 \cite{poh2011advancements}} & RGB & self-collected & 60 secs. & HR Estimation & RMSE = 5.63\% 
\\
\textbf{De Haan et al. 2013 \cite{Haan:13}} & RGB & self-collected & 2-30 secs. & HR Estimation & RMSE = 0.5 bpm \\
Tasli et al. 2014 \cite{landmarks2014RGB} & RGB & self-collected & 90 secs. & HR Estimation & MAE = 4.2 \%\\
Chen et al. 2014 \cite{hyperspectral2014stressdetection} & Hyperspectral & self-collected & 30-60 secs. & Stress Estimation & Qualitative\\
McDuff et al. 2014 \cite{mcduff2014multiband} & Multiband (RGBCO) & self-collected & 120 secs. & HR Estimation & Correlation = 1.0 \\
Chen et al. 2016 \cite{nir2016realsense} & RGB + NIR & self-collected & 90 secs. & HR Estimation & RMSE = 1.65\%\\
Li et al. 2016 \cite{li2016generalized} & RGB & 3DMAD and self-collected & 10 secs. & Face Antispoofing & EER = 4.71\% \\
\textbf{Wang et al. 2017 \cite{Wang:17}} & RGB & self-collected & 2-30 secs. & HR Estimation & SNR = 5.16 \\

Hernandez et al. 2018 \cite{hernandez2018time} & RGB and NIR & 3DMAD and self-collected & 1-60 secs. & HR Estimation & EER = 22\% (RGB) | 0\% (NIR) \\

\textbf{Chen et al. 2018 \cite{chen2018deepphys}} & RGB and NIR & HCI and self-collected & 10 secs. & Face Antispoofing & MAE = 4.57 bpm \\
\hline
\end{tabular}
}
\end{center}
\end{table*}

Classic rPPG works, usually called hand-crafted rPPG, use signal processing techniques for analyzing the images, and looking for slight color and illumination changes related with the BVP. More recent approaches use learning-based alternatives in order to train models capable of estimating the heart rate when receiving a face video in their input.  A selection of works related to remote photoplethysmography is shown in Table \ref{table1}.

One of the drawbacks of heart rate estimation using rPPG, compared to contact-based PPG techniques, is that rPPG is very sensitive to variability in the acquisition scenario. Some sources of this variability are \cite{hernandez2019quality}: changes in the illumination, blur, movement of the subject, vibration of the camera, resolution, frame rate, etc. Under constrained conditions, hand-crafted methods are able to achieve high accuracy levels, but in unconstrained scenarios learning-based approaches appear to dominate. 

The target of the present work is to perform a comparative study of state-of-the-art rPPG methods for heart rate estimation. We compare different methods presented in the literature, both hand-crafted and learning-based, tested against the same selection of videos from the COHFACE database \cite{heusch2017reproducible}. We check the performance of both types of approaches, identifying their respective strengths and weaknesses.

The rest of the paper is structured as follows. In Sect.~\ref{sec:system} we describe the database, the methods, and the experimental protocol. The results of the experiments are reported in Sect.~\ref{sec:results}. Conclusions and future work are finally drawn in Sect.~\ref{sec:conclusion}.

\section{EXPERIMENTAL FRAMEWORK}
\label{sec:system}


\subsection{Database}
\label{subsec:database}

\begin{figure}[t]
\includegraphics[width=\columnwidth]{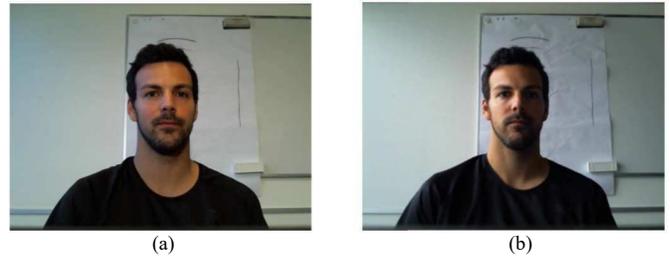}
\caption{Example images from the COHFACE database for its two illumination scenarios: (a) controlled illumination, and (b) uncontrolled illumination.}
\label{fig:database}
\end{figure}

\begin{figure*}[t]
\begin{center}
\includegraphics[width=0.9\textwidth]{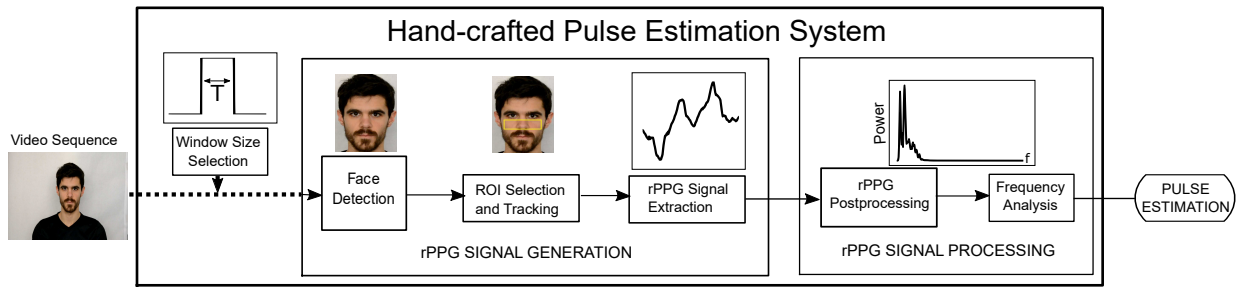}
\caption{Example of the workflow of a hand-crafted rPPG system for heart rate estimation using face videos. First, the different channels of the videos are processed to obtain raw rPPG signals. After that, the signals are postprocessed in order to extract a value for the heart rate for each temporal window.}
\label{fig:handcrafted}
\end{center}
\end{figure*}

The database used for the experiments is COHFACE \cite{heusch2017reproducible}. This database is composed of $160$ videos accompanied by the corresponding groundtruth heart rates of the subjects. In total $40$ people participated and $4$ videos were obtained from each of them. The duration of the recorded videos is $1$ minute each. The physiological signals relative to the heart rate were obtained by contact photoplethysmography (PPG). Both signals (video and PPG) are synchronized. The video sequences have a resolution of $640\times480$ pixels and were captured with a sampling frequency of $20$ frames per second. The heart rate signal was acquired at $256$ Hz. Videos were recorded in a laboratory. Regarding illumination conditions, for each subject two videos were obtained with controlled lighting, and other two with natural lighting. Examples of the database are shown in Fig.~\ref{fig:database}.

We decided to use this database for the evaluation because of several reasons. First, it is one of the few publicly available datasets oriented to heart rate extraction. It also presents enough amount of data for being able to train a learning-based model from scratch. Finally, it also presents a good balance between favorable and challenging conditions with respect to frame rate, resolution and illumination.

\subsection{Methods under Evaluation}
\label{subsec:systems}

The methods compared in this paper can be split into two different categories: hand-crafted and learning-based. 

\textbf{Hand-crafted methods} apply signal processing techniques to the video frames acquired by the camera sensors. These techniques analyze the video sequences and extract predefined features from the frames in order to estimate the heart rate signal. Hand-crafted methods have a major limitation: they may fail to model properly the highly nonlinear processes that may be occurring in the phenomenon they are trying to measure, as it happens in most of the processes related to human physiology, e.g. heart rate and respiration. An example of the typical workflow of a hand-crafted heart rate estimation method can be seen in Fig.~\ref{fig:handcrafted}.

The \textit{$4$ hand-crafted methods} that we compare in this work are: \cite{Verkruysse:08}, which emphasizes the role of luminance for obtaining physiological information; \cite{poh2011advancements}, which uses Blind Source Separation through ICA; \cite{Haan:13}, which focuses on chrominance as the source of information; and \cite{Wang:17}, which uses what the authors call The Plane Orthogonal to Skin$-$Tone (a linear function applied to the input data). Public open source implementations have been used for these $4$ hand-crafted methods \cite{mcduff2019iphys}.

As a representative example that illustrates typical hand-crafted methods, we now sketch the method \cite{poh2011advancements}. In this case the authors try to estimate the heart rate from a RGB video using Blind Source Separation (BSS) based on Independent Component Analysis (ICA), see Fig.~\ref{fig:handcrafted}:

\begin{itemize}
    \item A facial Region Of Interest (ROI) is obtained for each frame using automatic face tracking. The ROI is separated into its three RGB channels and the pixels of each channel are spatially averaged to obtain their mean values in each frame. This way the raw rPPG signals $x_1(t)$, $x_2(t)$ and $x_3(t)$ are generated by concatenating the mean values of all the processed frames. Subsequent processing is carried out using temporal windows of $30$ seconds with a window overlap of the $96.7$\% (increments of $1$ second). Normalization is applied to the raw rPPG signals in order to obtain signals with zero mean and unitary standard deviation.
    
    \item ICA is used to break down the normalized signals into three independent sources. Although the components obtained from ICA are not ordered, typically the second component contains a strong plethysmographic signal. Finally, the Fast Fourier Transform (FFT) is applied to the selected component in order to obtain its representation in the frequency domain. The final estimation of the heart rate will be the frequency component that corresponds to the maximum peak of the spectrum inside the band of frequencies considered physiologically possible for human heart rate.

\end{itemize}

\begin{table}[t]
\caption{Results of the comparison of heart rate estimation methods from face videos. MAE is computed for $3$ different video lengths: $60$, $30$, and $15$ seconds. Highlighted in bold are the best results obtained here for each video duration. In italic are the results published by Chen et al. 2018 \cite{chen2018deepphys}.
}
\label{results}
\begin{center}
\resizebox{\columnwidth}{!}{
\begin{tabular}{|c|c|c|c|c|}
\hline
\textbf{MAE (bpm)} & \textbf{60 s} & \textbf{30 s} & \textbf{15 s} & \textbf{Mean} \\
\hline\hline
Verkruysse et al. 2008 \cite{Verkruysse:08} & 12.86 & 14.30 & 16.91 & 14.69\\
Poh et al. 2011 \cite{poh2011advancements} & 16.51 & 13.75 & 14.90 & 15.05\\
De Haan et al. 2013 \cite{Haan:13} & 12.76 & 12.61 & 14.51 & 13.29\\
Wang et al. 2017 \cite{Wang:17} & 24.40 & 23.38 & 24.81 & 24.20\\
\textbf{Chen et al. 2018 \cite{chen2018deepphys} (COHFACE)} & \textbf{4.20} & \textbf{3.79} & \textbf{5.71} & \textbf{4.57}\\
\textit{Chen et al. 2018 \cite{chen2018deepphys} (HCI)} & \textit{-} & \textit{4.57} & \textit{-} & \textit{4.57} \\
\hline
\end{tabular}
}
\end{center}
\end{table}

On the other hand, \textbf{learning-based methods} for heart rate estimation are of recent development. In our experimental comparison we include the \textit{DeepPhys model} \cite{chen2018deepphys}. The main limitation of hand-crafted methods (i.e. its limited ability to model complex nonlinear functions) motivated the authors in \cite{chen2018deepphys} for designing a deep learning model that uses a Convolutional Attention Network (CAN) to perform physiological measurements from videos. This model achieves better results than traditional techniques since it is able of adapting its parameters to the non-linear nature of the heart rate signals. In addition, DeepPhys also permits the spatio-temporal visualization of physiological signals from RGB videos, e.g. the blood perfusion under the face skin.


\subsection{Experimental Protocol}
\label{subsec:protocol}

We have split the COHFACE database into train, development, and test subsets. Of the $40$ subjects in the database, $30$ have been used for training, $1$ has been used for validation and the remaining $9$ have been used for testing. We had $120$ minutes of video for training, $4$ minutes for validation, and $36$ minutes for testing. 


First, for the hand-crafted methods, as they don't need training, the performance and error measures are directly obtained from the test subset of COHFACE. The learning-based model, i.e. DeepPhys, needs to be trained to learn how to estimate the heart rate from video sequences. The original model pretrained in \cite{chen2018deepphys} is not public, so we have replicated its architecture and we have trained it from scratch with the training and validation partitions of COHFACE.

Tests have been performed for video sequences of $15$, $30$, and $60$ seconds of duration, and a sliding step of $1$ second, in order to evaluate the possible application of the methods when different amounts of data are available. The estimated and the groundtruth heart rates have been compared in beats per minute (bmp). The metric chosen to check the performance of the models is the Mean Absolute Error (MAE), which evaluates the distance between the estimated heart rate obtained and the groundtruth measure. 


\section{RESULTS}
\label{sec:results}

\subsection{Average Performance Comparison}

Table \ref{results} shows the results obtained with the hand-crafted and the learning-based approaches. It can be seen that for each method there are slight differences depending on the duration of the test videos. The mean MAE across the three lengths of the test videos is also presented. 

Among the hand-crafted methods, the first three are performing similarly with mean MAE values around $15$ bpm. However, the last one is obtaining worse error rates (around $24$ bpm mean MAE). In that paper, the authors presented an alternative rPPG method where they defined a projection Plane Orthogonal to the Skin$-$Tone (POS). As they wanted to keep the algorithm as simple as possible, they did not use the common band-pass filtering usually employed in other rPPG methods. Due to its characteristics, the use of the POS algorithm is interesting in those cases with favorable conditions, but it is more vulnerable to external factors like a heterogeneous illumination spectra, i.e. when two or more illumination sources are present. The COHFACE videos used for testing contain both controlled and uncontrolled illumination conditions, and this is likely the reason of the poor performance of the method from \cite{Wang:17} compared to the other hand-crafted alternatives. In general, the level of accuracy shown by the hand-crafted methods does not allow to apply them in many scenarios in which an error around $15$ bpm may be too high.

\begin{figure*}[t]
\begin{center}
\includegraphics[width=\linewidth]{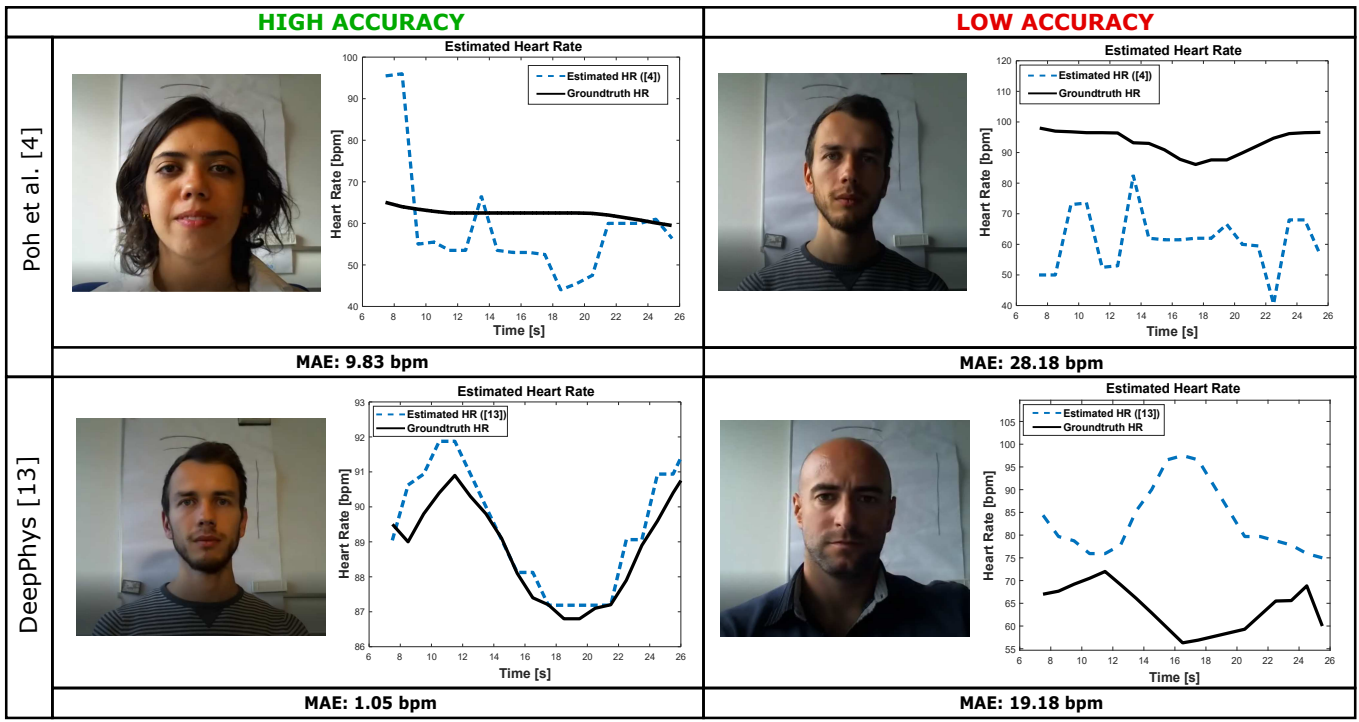}
\caption{\textbf{Accuracy for heart rate estimation} using a hand-crafted algorithm (Poh et al. \cite{poh2011advancements}) and our model of DeepPhys trained with COHFACE. MAE is computed for a video length of $15$ seconds, with a sliding step of $1$ second. We have selected the videos that obtain the highest and the lowest accuracy for both methods. The performance is higher (lower MAE) when dealing with videos with controlled illumination and low movement (left column). Results are worse when the illumination is natural and unconstrained (right column).}
\label{performance_worst_best}
\end{center}
\end{figure*}

\begin{figure*}[t]
\begin{center}
\includegraphics[width=\linewidth]{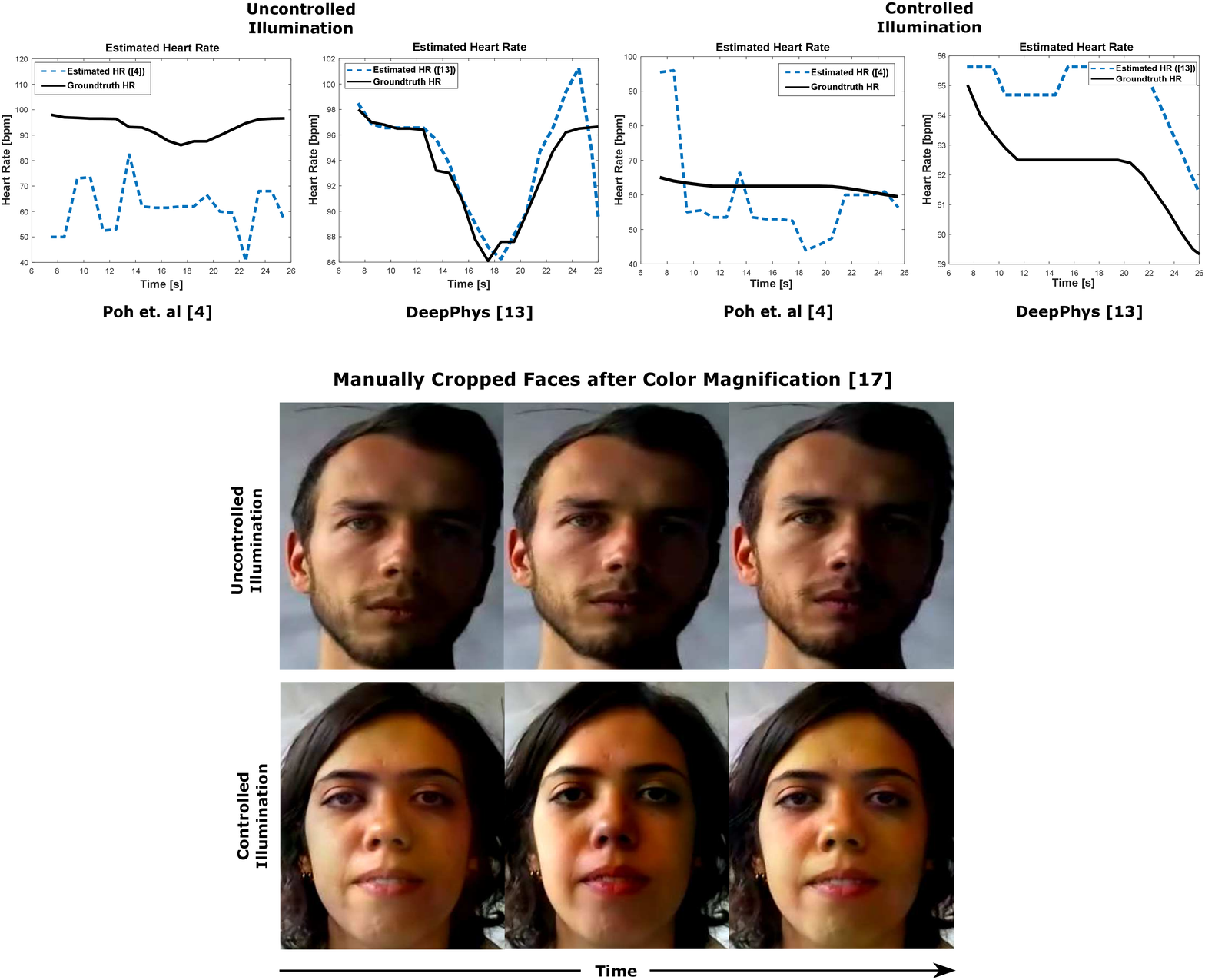}
\caption{\textbf{Examples of heart rate estimation.} The top row presents the estimation vs. the groundtruth for two videos, one with uncontrolled illumination (that obtained low accuracy with Poh et al. algorithm~\cite{poh2011advancements}) and other with controlled illumination (that achieved high accuracy with Poh's algorithm). For both cases we show the heart rate estimation using \cite{poh2011advancements} and also our DeepPhys \cite{chen2018deepphys} model implementation trained with COHFACE. HR is computed for a video length of $15$ seconds, with a sliding step of $1$ second. The second and third rows show some frames from both videos, whose color has been magnified using \cite{Wu12Eulerian} for showing how the pulse can be extracted more clearly in the case of controlled illumination acquisition.}
\label{magnified}
\end{center}
\end{figure*}


The results in Table \ref{results} also show that the DeepPhys model performs significantly better than the hand-crafted alternatives. This large difference in the performance is mainly caused by the ability of DeepPhys to apply the knowledge acquired from the training data, adapting its parameters to the conditions of each particular scenario. The learning-based model is capable of improving the hand-crafted results by around a $66$\% (from $15$ bpm to $5$ bpm of mean MAE). This is caused by its ability to capture the non-linear nature of the physiological data. In addition, the hand-crafted models have been designed to operate under certain conditions of illumination, resolution, etc, while the learning process allows to adapt the DeepPhys model to each specific scenario. On the other hand, the drawback of the learning-based approaches is that a high amount of data is needed in order to train a deep model from scratch, while hand-crafted methods do not need that training stage, being able to work directly on the test data. However, it is possible to reduce the necessary amount of training data using a model that has been previously trained with a similar dataset. That model can be then fine-tuned with few additional data to get adapted to the new scenario.

Regarding the video duration, increasing the length above $30$ seconds does not improve the accuracy of the methods significantly. With video segments of at least $15$ seconds the rPPG algorithms achieve their best results, so in case of wanting to apply this technology to a real scenario it seems that it will not be necessary to process longer videos.

We should also mention that in \cite{chen2018deepphys} the authors trained their DeepPhys model with videos from the Mahnob-HCI database \cite{mahnobhci} with a duration of $30$ seconds, achieving a MAE value of $4.57$ bpm, close to the performance obtained on average with the data we used in our experiments. It seems that the Convolutional Atention Network of DeepPhys is able to perform robustly even when facing data with different levels of illumination, resolution, and frame rate.

\subsection{Comparison across Time on Selected Videos}

Here we compare one of the best performing hand-crafted algorithms, Poh et al. \cite{poh2011advancements}, with the DeepPhys model trained in this paper, using a window length of $15$ seconds and a sliding step of $1$ second. In this setup, we selected the video with the highest accuracy and the one with the lowest for each method. The accuracy values and some frames of these videos are shown in Fig.~\ref{performance_worst_best}.

We have found that both methods obtain inaccurate results when the videos contain a high amount of movement, when the person blinks with excessive strength, or when he/she does any type of unwitting movement. For both methods, the videos with the best accuracy for heart rate estimation belong to the controlled illumination scenario. On the other hand, the worst results are obtained in videos with uncontrolled illumination. Hand-crafted methods like Poh et al. \cite{poh2011advancements} look for changes in the skin tone at some Regions Of Interest (ROI) inside the face. Due to the unconstrained illumination conditions at these regions, e.g. one of the cheeks in the images shown in Fig.~\ref{performance_worst_best} (right column), the ROI image levels can vary significantly across time according to the varying illumination, degrading the raw rPPG signal and making the accuracy to drop significantly.  

In order to illustrate the relevance of the illumination conditions when using hand-crafted rPPG algorithms, we have selected a pair of videos from the testing dataset, one of them captured under controlled illumination conditions and the other with uncontrolled illumination, and we have magnified their color using Eulerian Video Magnification \cite{Wu12Eulerian}, a method that highlights the color changes across the frames. These changes may be caused by the pulse, the external illumination variability, or both. Fig.~\ref{magnified} shows how the color changes due to human pulse are easier to see in the controlled illumination scenario. This explains why hand-crafted rPPG methods such as Poh et al. \cite{poh2011advancements} obtain higher accuracy under these acquisition conditions. 






According to the general results shown in Table \ref{results}, the performance is significantly higher when using DeepPhys compared to the hand-crafted methods. However, there are also cases in which the deep learning algorithm obtains inaccurate results. Taking a look to the MAE obtained for each individual video, it seems that the drop in the accuracy of the deep learning approach is not so intimately related with the illumination conditions, but with other factors like if the users are wearing complements (e.g. glasses) or their skin tone. The training process has made the CNN more robust to illumination variability, since the training set contains the same number of videos with controlled and uncontrolled illumination conditions. This made possible for the network to learn how to extract a precise heart rate estimation in both cases. However, the number of samples of users with dark skin tones or wearing glasses is not so frequent in the training set, so DeepPhys did not learn so accurately how to deal with these conditions. In order to improve the accuracy of DeepPhys it will be interesting to feed the network with a dataset that contains a higher number of these types of samples.

Summarizing, learning-based models have the potential of achieving accuracy values much higher than the ones from hand-crafted approaches, as far as they are properly trained with enough representative data. These methods are also more robust to external factors such as low illumination or resolution. On the other hand, their main drawback is the need for a high amount of data for training the models, which are not always available.


\section{CONCLUSIONS AND FUTURE WORK}
\label{sec:conclusion}



In this paper we performed a comparative evaluation of heart rate estimation methods using face videos. This study has taken into account both hand-crafted and learning-based state-of-the-art methods. Results have been provided using the COHFACE database, which is publicly available \cite{heusch2017reproducible}.

For the evaluation we have selected $4$ different hand-crafted methods from the most relevant in the literature. Each one of these methods emphasizes a particular aspect from the vector space generated by the camera sensors. All of these methods have obtained a fairly similar performance. 


Under our test scenario conditions, the hand-crafted methods have found a performance barrier around $15$ bpm MAE. Human physiology turns out to be complex and highly non-linear and this may be the reason why hand-crafted methods fail to achieve better results. 

On the other hand, the suitability of the learning-based models for estimating heart rate measurements through rPPG has been verified. Our DeepPhys implementation has obtained better results than the hand-crafted methods, reaching an error around $5$ bpm MAE, similar to the experiments reported by Chen and McDuff in \cite{chen2018deepphys}, even though in the present work, the COHFACE database contained videos with less controlled illumination than the ones used in the original work, making us think that this type of models could be applied to even more challenging scenarios. The deep neural network is capable of capturing non-linearity thanks to the training process. This result encourages the research in learning-based methods for heart rate detection using face videos, and also for studying other physiological signals that may present a non-linear behavior \cite{hernandez2019edbb}. 

For future work it could be interesting to test learning-based methods with input data acquired in less controlled conditions. This could be done by increasing the distance between the camera and the subject, introducing a higher amount of movement in the videos, making illumination conditions even less controlled, etc. The objective is to develop a model capable of accurately estimating a person's heart rate through rPPG from a recording made in the wild \cite{2018_TIFS_SoftWildAnno_Sosa}. Fine-tuning pre-trained models for adapting them to new scenarios is also an alternative that may be explored, in order to surpass the lack of training data in some cases. Using other parts of the body apart from the face for rPPG is another possibility, e.g. the skin of the torso, the arms, or the legs. This could help to make the heart rate estimation more robust in the cases where the face data is of low quality \cite{alonso2011quality} or directly non available.

\section{ACKNOWLEDGMENTS}
\label{sec:ack}

This work has been supported by projects: IDEA-FAST
(IMI2-2018-15-two-stage-853981), PRIMA
(ITN-2019-860315), TRESPASS-ETN
(ITN-2019-860813), BIBECA (RTI2018-
101248-B-I00 MINECO/FEDER), and edBB (UAM). J.H.O. is supported by a PhD Scholarship from UAM.
Portions of the research in this paper used the COHFACE Dataset made available by the Idiap Research Institute, Martigny, Switzerland.

\bibliographystyle{IEEEbib}
\bibliography{egbib}

\begin{thebibliography}{10}

\bibitem{allen2007photoplethysmography}
John Allen,
\newblock ``Photoplethysmography and its application in clinical physiological
  measurement,''
\newblock {\em Physiological Measurement}, vol. 28, no. 3, 2007.

\bibitem{thermal2007garbey}
Marc Garbey, Nanfei Sun, Arcangelo Merla, and Ioannis Pavlidis,
\newblock ``Contact-free measurement of cardiac pulse based on the analysis of
  thermal imagery,''
\newblock {\em IEEE {T}ransactions on {B}iomedical {E}ngineering}, vol. 54, no.
  8, pp. 1418--1426, 2007.

\bibitem{Verkruysse:08}
Wim Verkruysse, Lars~O Svaasand, and J~Stuart Nelson,
\newblock ``Remote plethysmographic imaging using ambient light.,''
\newblock {\em Optics Express}, vol. 16, no. 26, pp. 21434--21445, 2008.

\bibitem{poh2011advancements}
Ming-Zher Poh, Daniel~J McDuff, and Rosalind~W Picard,
\newblock ``Advancements in noncontact, multiparameter physiological
  measurements using a webcam,''
\newblock {\em IEEE {T}ransactions on {B}iomedical {E}ngineering}, vol. 58, no.
  1, pp. 7--11, 2011.

\bibitem{Haan:13}
Gerard {de Haan} and Vincent {Jeanne},
\newblock ``Robust pulse rate from chrominance-based r{PPG},''
\newblock {\em IEEE {T}ransactions on {B}iomedical {E}ngineering}, vol. 60, no.
  10, pp. 2878--2886, 2013.

\bibitem{landmarks2014RGB}
H~Emrah Tasli, Amogh Gudi, and Marten den Uyl,
\newblock ``Remote {PPG} based vital sign measurement using adaptive facial
  regions,''
\newblock in {\em Proc. {IEEE} {I}nternational {C}onf. on {I}mage {P}rocessing
  ({ICIP})}, 2014, pp. 1410--1414.

\bibitem{hyperspectral2014stressdetection}
Tong Chen, Peter Yuen, Mark Richardson, Guangyuan Liu, and Zhishun She,
\newblock ``Detection of psychological stress using a hyperspectral imaging
  technique,''
\newblock {\em IEEE {T}ransactions on {A}ffective {C}omputing}, vol. 5, no. 4,
  pp. 391--405, 2014.

\bibitem{mcduff2014multiband}
Daniel McDuff, Sarah Gontarek, and Rosalind~W Picard,
\newblock ``Improvements in remote cardiopulmonary measurement using a five
  band digital camera,''
\newblock {\em IEEE {T}ransactions on {B}iomedical {E}ngineering}, vol. 61, no.
  10, pp. 2593--2601, 2014.

\bibitem{nir2016realsense}
Jie Chen, Zhuoqing Chang, Qiang Qiu, Xiaobai Li, Guillermo Sapiro, Alex
  Bronstein, and Matti Pietik{\"a}inen,
\newblock ``Real{S}ense = real heart rate: {I}llumination invariant heart rate
  estimation from videos,''
\newblock in {\em Proc. {I}nternational {C}onference on {I}mage {P}rocessing
  {T}heory {T}ools and {A}pplications ({IPTA})}, 2016.

\bibitem{li2016generalized}
Xiaobai Li, Jukka Komulainen, Guoying Zhao, Pong-Chi Yuen, and Matti
  Pietik{\"a}inen,
\newblock ``Generalized face anti-spoofing by detecting pulse from face
  videos,''
\newblock in {\em Proc. {I}nternational {C}onference on {P}attern {R}ecognition
  ({ICPR})}, 2016, pp. 4244--4249.

\bibitem{Wang:17}
Wenjin {Wang}, Albertus~C. {den Brinker}, Sander {Stuijk}, and Gerard {de
  Haan},
\newblock ``Algorithmic principles of remote {PPG},''
\newblock {\em IEEE {T}ransactions on {B}iomedical {E}ngineering}, vol. 64, no.
  7, pp. 1479--1491, 2017.

\bibitem{hernandez2018time}
Javier Hernandez-Ortega, Julian Fierrez, Aythami Morales, and Pedro Tome,
\newblock ``Time analysis of pulse-based face anti-spoofing in visible and
  {NIR},''
\newblock in {\em {IEEE} {C}onference on {C}omputer {V}ision and {P}attern
  {R}ecognition {W}orkshops}, 2018, pp. 544--552.

\bibitem{chen2018deepphys}
Weixuan Chen and Daniel McDuff,
\newblock ``Deep{P}hys: {V}ideo-based physiological measurement using
  convolutional attention networks,''
\newblock in {\em Proc. of the European Conf. on Computer Vision (ECCV)}, 2018,
  pp. 349--365.

\bibitem{hernandez2019quality}
Javier Hernandez-Ortega, Shigenori Nagae, Julian Fierrez, and Aythami Morales,
\newblock ``Quality-based pulse estimation from {NIR} face video with
  application to driver monitoring,''
\newblock in {\em Proc. {I}berian {C}onference on {P}attern {R}ecognition and
  {I}mage {A}nalysis}. Springer, 2019, pp. 108--119.

\bibitem{heusch2017reproducible}
Guillaume Heusch, Andre Anjos, and Sebastien Marcel,
\newblock ``A reproducible study on remote heart rate measurement,''
\newblock {\em arXiv preprint arXiv:1709.00962}, 2017.

\bibitem{mcduff2019iphys}
Daniel McDuff and Ethan Blackford,
\newblock ``i{P}hys: {A}n open non-contact imaging-based physiological
  measurement toolbox,''
\newblock in {\em Proc. {I}nternational {C}onference of the {IEEE}
  {E}ngineering in {M}edicine and {B}iology {S}ociety ({EMBC})}, 2019, pp.
  6521--6524.

\bibitem{Wu12Eulerian}
Hao-Yu Wu, Michael Rubinstein, Eugene Shih, John Guttag, Fr\'{e}do Durand, and
  William~T. Freeman,
\newblock ``Eulerian {V}ideo {M}agnification for {R}evealing {S}ubtle {C}hanges
  in the {W}orld,''
\newblock {\em {ACM} {T}ransactions on {G}raphics}, vol. 31, no. 4, 2012.

\bibitem{mahnobhci}
Mohammad {Soleymani}, Jeroen {Lichtenauer}, Thierry {Pun}, and Maja {Pantic},
\newblock ``A multimodal database for affect recognition and implicit
  tagging,''
\newblock {\em IEEE {T}ransactions on {A}ffective {C}omputing}, vol. 3, no. 1,
  pp. 42--55, 2012.

\bibitem{hernandez2019edbb}
Javier Hernandez-Ortega, Roberto Daza, Aythami Morales, Julian Fierrez, and
  Javier Ortega-Garcia,
\newblock ``ed{BB}: {B}iometrics and behavior for assessing remote education,''
\newblock in {\em Workshop on {A}rtificial {I}ntelligence for {E}ducation, AAAI
  Conference on Artificial Intelligence}, 2020.

\bibitem{2018_TIFS_SoftWildAnno_Sosa}
Ester Gonzalez-Sosa, Julian Fierrez, Ruben Vera-Rodriguez, and Fernando
  Alonso-Fernandez,
\newblock ``Facial soft biometrics for recognition in the wild: Recent works,
  annotation and {COTS} evaluation,''
\newblock {\em IEEE {T}rans. on {I}nformation {F}orensics and {S}ecurity}, vol.
  13, no. 8, pp. 2001--2014, 2018.

\bibitem{alonso2011quality}
Fernando Alonso-Fernandez, Julian Fierrez, and Javier Ortega-Garcia,
\newblock ``Quality measures in biometric systems,''
\newblock {\em IEEE {S}ecurity \& {P}rivacy}, vol. 10, no. 6, pp. 52--62, 2011.

\end{thebibliography}

\end{document}